\documentclass[lettersize,journal]{IEEEtran}
\usepackage{amsmath,amsfonts}
\usepackage{algorithmic}
\usepackage{array}
\usepackage[caption=false,font=normalsize,labelfont=sf,textfont=sf]{subfig}
\usepackage{textcomp}
\usepackage{stfloats}
\usepackage{url}
\usepackage{verbatim}
\usepackage{graphicx}
\usepackage{epsfig} 
\usepackage{times} 
\usepackage{amssymb}  
\usepackage{booktabs}
\usepackage{multirow}
\usepackage[table,xcdraw]{xcolor}
\graphicspath{ {./images/} }
\usepackage{flushend}
\usepackage{float}
\usepackage{siunitx}
\usepackage{ulem}
\def\BibTeX{{\rm B\kern-.05em{\sc i\kern-.025em b}\kern-.08em
    T\kern-.1667em\lower.7ex\hbox{E}\kern-.125emX}}
\usepackage{balance}
\begin{document}
\title{Experience-Learning Inspired Two-Step Reward Method for Efficient Legged Locomotion Learning Towards Natural and Robust Gaits
}
\author{Yinghui Li, Jinze Wu, Xin Liu, Weizhong Guo$^{*}$, Yufei Xue  
\thanks{All authors are with School of Mechanical Engineering, Shanghai Jiao Tong University, Shanghai, China. *Corresponding author}
\thanks{This work was supported by the National Key Research and Development Plan(2021YFF0307901).}}

\maketitle

\begin{abstract}
Multi-legged robots offer enhanced stability in complex terrains, yet autonomously learning natural and robust motions in such environments remains challenging. Drawing inspiration from animals' progressive learning patterns, from simple to complex tasks, we introduce a universal two-stage learning framework with two-step reward setting based on self-acquired experience, which efficiently enables legged robots to incrementally learn natural and robust movements. In the first stage, robots learn through gait-related rewards to track velocity on flat terrain, acquiring natural, robust movements and generating effective motion experience data. In the second stage, mirroring animal learning from existing experiences, robots learn to navigate challenging terrains with natural and robust movements using adversarial imitation learning. To demonstrate our method's efficacy, we trained both quadruped robots and a hexapod robot, and the policy were successfully transferred to a physical quadruped robot GO1, which exhibited natural gait patterns and remarkable robustness in various terrains.
\end{abstract}


\section{Introduction}
\IEEEPARstart{T}{he} intersection of biology and robotics has been a fertile ground for mutual learning and advancements\cite{ramdya2023neuromechanics}. Robotics experts aspire to learn from biological principles to design robots capable of robust movement in complex environments, but the realization of such designs remains a challenge. While roboticists have been inspired by biological structures to develop various legged robots, existing research has not yet succeeded in replicating the rapid learning and acquisition of natural, robust movement in complex environments as seen in biological counterparts. This has led to extensive research focused on understanding potential biological motion mechanisms, with the aim of efficiently analyzing, validating, and incorporating them into robotic systems.

Animal locomotion learning typically progresses from simple tasks, like gait learning on flat ground, to more complex movements in varied terrains, developing natural and robust motion habits. However, in this progressive learning model, how previously acquired motion experiences influence the learning of new complex movements, and the underlying logic of this biological subconscious learning, remains unknown. This paper suggests that the learned experience from previous tasks could act as induced reward signaling to efficiently aid in mastering complex locomotion for legged robots, potentially revealing key aspects of biological motion learning and furthering research in robust control for robots.

\begin{figure}[tbp]
\includegraphics[width=\linewidth]{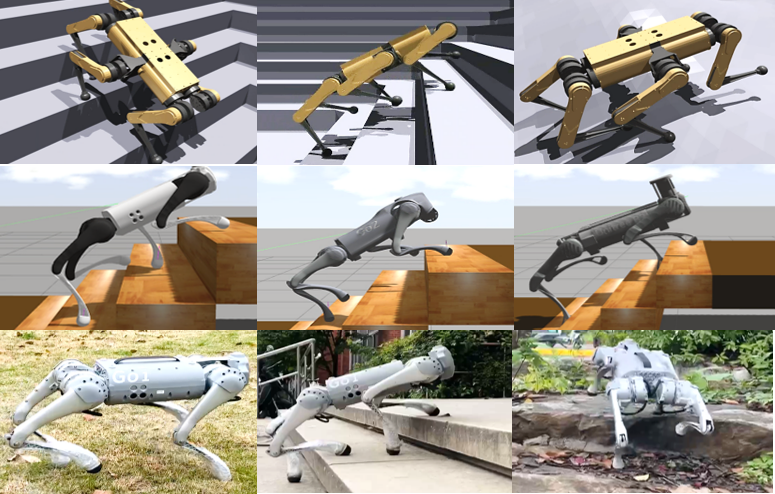}
\centering
\caption {Our approach develops a hardware-robust policy, equipping legged robots with neural network control to achieve stable and naturally robust gaits across diverse terrains. In the top part of our testing, hexapod and quadruped robots like HEX, Unitree-Go1, Go2, and B2 showcase the effectiveness of our trained controllers in producing natural, robust diagonal gaits, even in challenging settings like staircases. In the bottom part, we validate the transferability of our training results by successfully applying the trained strategies to the real robot Go1, exemplifying our method's practical applicability.}
\label{fig:locomotion}
\end{figure}

In current research, reinforcement learning method plays a crucial role in the locomotion learning of legged robots, enabling them to traverse complex environments effectively. However, current research often struggles to generate natural and robust movement patterns in complex environments solely through reward functions. Additionally, the learning process tends to be isolated, with different setups and reward functions required for various tasks, making it challenging to effectively leverage experiences across different tasks. These limitations contrast with biology's progressive learning, where organisms use prior experiences to adeptly master complex tasks, swiftly developing natural and robust movements. Addressing how to integrate this biological-style learning into the strategy training process becomes crucial, potentially revolutionizing the existing research paradigm.

In this letter, we introduce a novel bioinspired two-stage learning framework with two-step reward setting that leverages prior motion experiences from simple locomotion tasks, utilizing reinforcement learning algorithms and adversarial imitation learning method to effectively induce naturally robust motion behaviors in complex terrains. This method has been successfully applied to both quadruped and hexapod robots, allowing them to achieve natural and robust diagonal gaits in challenging environments.

The main contributions are listed as follows:
\begin{enumerate}
\item We introduce a two-stage learning framework with efficient reward method that utilizes the self prior motion experience to facilitate their efficient mastery of naturally robust locomotion in complex environments. 
\item Specific rewards setting and training for different robots are developed, demonstrating efficient application and validation of the proposed methods.
\item Employing a Teacher-Student strategy, these learned methods are successfully implemented on real robots Go1, showcasing their capability to execute natural and robust locomotion in various challenging environments.
\end{enumerate}  

\section{Related Work}
\subsection{Bio-inspired Progressive Learning Patterns}
\subsubsection{Zoology Progressive Learning Patterns}
\noindent Living beings exhibit a progressive learning pattern where complex tasks are autonomously broken down into simpler sub-tasks, such as juxtaposed, concatenated, and concurrent tasks, ultimately culminating in the comprehensive completion of the complex task. For instance, human infants and newborn mammals\cite{dominici2011locomotor}, like pigs\cite{babypiglearntowalk}, first learn to walk on flat ground, starting with basic balance, then progressing to standing, simple steps, and eventually smooth walking. This step-by-step learning approach, where the motion experience gained in initial stages significantly influences the learning of more complex tasks, leads to naturally robust locomotion in complex environments.

\subsubsection{Robotics Applications of Progressive Learning Patterns}
\noindent Robotic control algorithms commonly use progressive methods to simplify and then reintegrate complex tasks.  To achieve robust velocity following movement of quadruped robots in challenging environments, researchers\cite{hwangbo2019learning}\cite{shi2022reinforcement} first establish basic reference trajectories through optimization for tasks like flat ground navigation, and then enhance these foundations with learning methods tailored to complex tasks, ensuring adaptability and effectiveness across diverse settings.To accomplish complex integration tasks, another researcher\cite{yang2020multi}\cite{thor2022versatile} broke down the task into multiple sub-tasks for individual learning, later combining these results to effectively realize the overarching complex task. These methods often rely heavily on the designer's preconceptions, such as predefined reference trajectories and task decomposition methods. Such reliance can significantly limit the outcomes of learning, diverging from the way organisms learn through self experience. 

\subsection{Learning method for Locomotion}

\subsubsection{Reinforcement Learning for Locomotion}

\noindent Data-driven algorithms, notably Reinforcement Learning (RL), have been increasingly used in recent years for controlling legged robots\cite{hwangbo2019learning}\cite{lee2020learning}\cite{miki2022learning}. The neural network controllers, trained by reinforcement learning algorithms, have enabled robust locomotion in legged robots. However, fulfilling natural, stable, and other movement requirements for legged robots, particularly in complex terrains, remains a challenge when relying solely on manually set reward functions for learning methods.

\subsubsection{Motion Imitation Learning}
\noindent Designing effective reward functions for legged robots in Reinforcement Learning to elicit desired behaviors from an agent remains a significant challenge. One approach  \cite{peng2018deepmimic} \cite{peng2020learning} to enhance the quality of learning is through the imitation of animal motion capture or hand-authored animation data. This strategy, while effective for replicating individual motion clips, faces challenges in imitating multiple reference motions with a single phase variable. Addressing this, \cite{peng2021amp} introduced Adversarial Motion Priors (AMP), which applies the GAIL framework\cite{ho2016generative} to discern whether a state transition \(\left(s_t,s_{t+1}\right)\) is authentically from the data set or fabricated by the agent. This method allows simulated agents to execute complex tasks while adopting motion styles from extensive, unstructured motion data sets, and has been widely implemented in legged robots. \cite{escontrela2022adversarial} \cite{vollenweider2022advanced}, Current imitation learning sources, typically derived from animals or pre-modeling methods\cite{wjz}, struggle with adapting to robots with varying configurations like parallel or elastic legs. Conversely, organisms naturally bypass such scale and configuration constraints, learning robust behavior patterns by evolving from their existing motion experiences. 

In this letter, we advocate for a bionic two-stage progressive locomotion learning approach, aiming to emulate the progressive self-learning process observed in living beings, and to effectively induce naturally robust motion behaviors of legged robos in complex terrains. 

\section{Method}

\begin{figure*}[htbp]
\includegraphics[width=\linewidth]{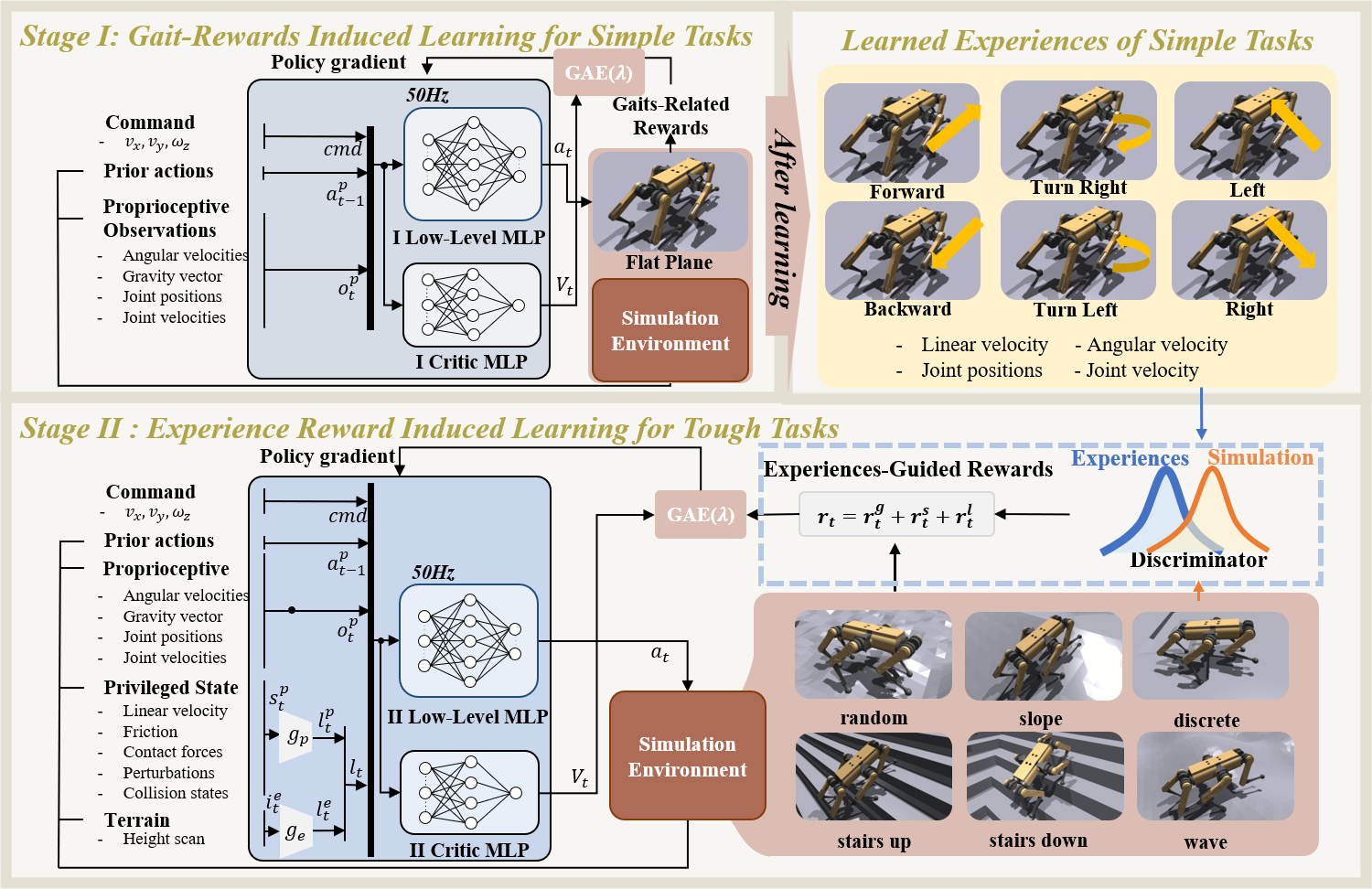}
\setlength{\abovecaptionskip}{-0cm}
\setlength{\belowcaptionskip}{-0cm}
\vspace{-0.cm}
\centering
\caption{Our method comprises two main stages: rewards-rewards induced learning for simple tasks and experience-reward induced learning for rough tasks, culminating in deployment on real robots using a teacher-student strategy. In the first stage, the robot is trained to track velocity commands with a diagonal gait in a flat terrain environment. We incorporate gait-related reward functions to effectively constrain the robot's gait, foot trajectory, and body state, enabling it to achieve a natural and robust diagonal gait. After training, the network generates motion state data specific to the task, storing experiences such as the robot's body state (linear and angular velocity) and joint states (position and velocity). In the second stage, the robot need track velocity commands with a diagonal gait in complex environments. Additional privileged information like terrain data, body linear velocity, and dynamic parameters are fed into the network as observations. The robot's previously acquired motion experiences serve as a reference, training a discriminator network to identify similarities between current tasks and past experiences, and to generate style reward signals. These are combined with task rewards and regularization rewards to update the actor and critic networks. During deployment, the teacher-student method is used to encode privileged information from proprioceptive sensing, facilitating successful implementation on real robots.}
\label{fig:method}
\end{figure*}

In this letter, the objective is to develop a locomotion controller capable of operating Legged robots without vision information that performs natural and robust movement. Our approach deconstructs this task into two components: gait learning in flat terrain and robust movement in complex terrains, culminating in real-world deployment using a teacher-student strategy. The overall methodology is illustrated in Fig. \ref{fig:method}, with the algorithm applied to both several quadruped and a hexapod robot. 

Characterized by high redundancy, the hexapod robot can maintain stability in complex environments, even with motor failures. This redundancy, while offering stability, creates a vast exploration space, challenging the definition of reward functions during training. Our paper primarily focuses on the hexapod to demonstrate naturally robust diagonal gait learning in such environments, showcasing our biologically inspired two-stage learning framework. And we utilise unitree go1 to test the hardware robustness of the trained controller.

\begin{table*}[htbp]
\label{tab:reward details}
\centering
\footnotesize
\caption{Reward terms for velocity commands tracking task, Regularization (Stability, smoothness, Safety), and Specific Stage.}
\begin{tabular}{@{}cccccc@{}}
\toprule
\textbf{Stage} & \multicolumn{2}{c}{\textbf{Term}}                           & \textbf{annotation}       & \textbf{equation}                              & \textbf{scale}\\ \midrule
{\multirow{12}{*}[-10pt]{\textbf{For Both}}} &
\multicolumn{2}{c}{\multirow{2}{*}{\textbf{Task} $r^g$}}    & Linear velocity tracking  & $\exp \left(\Vert\mathbf{v}_{t,xy}^{\rm des}-\mathbf{v}_{t,xy}\Vert_2 \big/ 0.15 \right)$        & 1$dt$\\
\multicolumn{3}{c}{}                                        & Angular velocity tracking & $\exp \left(\Vert{\omega}_{t,z}^{\rm des}-{\omega}_{t,z}\Vert_2 \big/ 0.15 \right)$               & 0.8$dt$\\ \cmidrule(l){2-6}
&\multirow{10}{*}{\textbf{Regularization} $r^l$} & \multirow{3}{*}{Stability}  & Linear velocity penalty   & $-{v}_{t,z}^2$                                                                     & 2$dt$\\
&&& Angular velocity penalty  & $-\Vert\boldsymbol{\omega}_{t,xy}\Vert_2$                                                           & 0.05$dt$\\
&&& Body height penalty    & $-\Vert\boldsymbol{h}_z-\boldsymbol{h}_z^{\text {des}}\Vert_2$  & 0.2$dt$\\ \cmidrule(l){3-6} 
&& \multirow{3}{*}{Smoothness}      & Joint torque              & $-\Vert\boldsymbol{\tau}\Vert_2$                                                                 & $1e^{-5}dt$\\
&&& Joint acceleration        & $-\Vert\mathbf{\ddot q}\Vert_2$                                                                     & $2.5e^{-7}dt$\\

&&& Action rate               & $-\Vert \mathbf{a}_{t-1}-\mathbf{a}_t\Vert_2$                                                       & 0.01$dt$\\ \cmidrule(l){3-6} 
&& \multirow{4}{*}{Safety}       & Collisions                & $-n_{collision}$                                                                                 & 0.1$dt$\\
&&& Joint torque limits       & $-\Vert\max\left(\left|\boldsymbol{\tau}_t \right|-\boldsymbol{\tau}^{limit},0\right)\Vert_2$      & 0.01$dt$\\
&&& Joint velocity limits     & $-\Vert\max\left(\left|\boldsymbol{\dot q}_t \right|-\boldsymbol{\dot q}^{limit},0\right)\Vert_2$  & 0.1$dt$\\
&&& Contact force penalty     & $-\Vert\max\left(\left|\mathbf{f}_t \right|-\mathbf{f}^{limit},0\right)\Vert_2$                    & 0.02$dt$\\ \midrule
{\multirow{4}{*}[-20pt]{{\textbf{Stage I}}}} 
& \multicolumn{2}{c}{\multirow{4}{*}[-20pt]{\textbf{Gait-related Rewards} $r^{gait}$}} 
& Swing phase tracking(force)    & $\sum_{\text {foot }}\left[1-C_{\text {foot }}^{\text {cmd }}\left(\boldsymbol{\theta}^{\mathrm{cmd}}, t\right)\right] \exp \left\{-\left|\mathbf{f}^{\text {toot }}\right|^2 / \sigma_{c f}\right\}$  & 4$dt$\\ \cmidrule(l){4-6} 
&&& Stance phase tracking(velocity)    & $\sum_{\text {foot }}\left[C_{\text {foot }}^{\text {cmd }}\left(\theta^{\text {cmd }}, t\right)\right] \exp \left\{-\left|\mathbf{v}_{x y}^{\text {foot }}\right|^2 / \sigma_{c v}\right\}$  & 4$dt$\\ \cmidrule(l){4-6} 
&&& Raibert footswing tracking    & $\left(\mathbf{p}_{x, y, \text { foot }}^f-\mathbf{p}_{x, y, \text { foot }}^f\left(\boldsymbol{s}_y^{\mathrm{des}}\right)\right)^2$  & 10$dt$\\ \cmidrule(l){4-6} 
&&& footswing height tracking    & $\sum_{\text {foot }}\left(\boldsymbol{h}_{z \text { foot }}^f-\boldsymbol{h}_z^{f, \mathrm{des}}\right)^2 C_{\mathrm{foot}}^{\mathrm{des}}\left(\boldsymbol{\theta}^{\mathrm{des}}, t\right)$  & 2$dt$\\ \midrule
{{\textbf{Stage II}}} & \multicolumn{2}{c}{\textbf{Style} $r^e$}                   & Score of discriminator    & $\max \left[0, 1-0.25\left(d_t^{\rm score}-1\right)^2 \right]$  & 1$dt$\\ \bottomrule
\end{tabular}
\end{table*}

\subsection{Reinforcement Learning Problem Formulation}
The proposed method to the control issue adopts a discrete-time dynamic model. At each discrete interval, denoted by time step $t$, the system's state is completely characterized by $\boldsymbol{x}_t$. An action $\boldsymbol{a}_t$ is executed according to the policy, leading to a progression to the subsequent state $\boldsymbol{x}_{t+1}$, which occurs with a probability defined by \(P\left({{\boldsymbol{x}_{t + 1}}\mid{\boldsymbol{x}_t},{\boldsymbol{a}_t}}\right)\), and yields a reward $r_t$. The objective within the realm of Reinforcement Learning (RL) is to identify the policy's optimal parameters \({\pi _\theta }\) that will optimize the cumulative expected return, taking into account the decay of future rewards as expressed by the discount factor $\gamma ^t$. This is mathematically represented as the maximization of the function:
\begin{equation} \label{eq:RLObjective}
J\left( \theta  \right) = {\mathbb{E}_{{\pi _\theta }}}\left[ {\sum\limits_{t = 0}^\infty  {{\gamma ^t}{r_t}} } \right]
\end{equation}

\textbf{Observation Space:}  The observation spaces differ between stages due to task differences. For flat terrain gait learning and velocity tracking, observations \(\boldsymbol{x}_t^{\text{I}}\) include proprioceptive data \(\boldsymbol{o}_t^p\) (body angular velocity, gravity vector, joint positions, velocities), velocity commands, and prior action commands. In complex environments, observations \(\boldsymbol{x}_t^{\text{II}}\) expand to include terrain height scan \(\boldsymbol{i}_t^e\) and privileged information  \(\boldsymbol{s}_t^p\) like body linear velocity and dynamic parameters (friction, contact forces, perturbations, collision states). Terrain information is derived from numerous points around the robot, indicating their vertical distance to the robot's base. To manage the complexity, terrain and privileged information are each encoded separately using multi-layer perceptron networks before being fed into a Low-Level MLP for inference. For deployment training, the policy state, \(\boldsymbol{x}_t^{\text{deploy}}\), is limited to proprioceptive observations \(\boldsymbol{o}_t^p\) only. 

\textbf{Action Space:} The policy action $\boldsymbol{a}_{t}$ is an 18-dimensional vector interpreted as a target joint position offset, which is added to the time-invariant nominal joint position to specify the target motor position for each joint.  These position targets would be used to compute desired torques by low-level joint PD controllers  $\tau=\boldsymbol{K}_p\left(\boldsymbol{q}_d-\right.q)+\boldsymbol{K}_d\left(\dot{q}_d-\dot{q}\right)$, in which we determine the target joint velocity to 0.

\textbf{Reward Design:} Across different stages, there are several consistent reward function settings: a task-focused reward \(r^g_t\) and a regularization reward \(r^l_t\). The task reward is designed to ensure accurate tracking of commanded velocities, while the regularization reward promotes stability, smoothness, and safety. This includes penalties for base instability and joint motion incoherence, alongside bonuses for stride duration. In addition to the consistent reward settings across stages, there are stage-specific adjustments: the first stage, focused on gait learning in flat plane, incorporates a specific gait reward function \(r^{gait}_t\). In the second stage, which centers on experience-guided natural and robust kinematic learning, a distinct reward function \(r^e_t\) is implemented. The tripod-style reward, based on adversarial motion priors, motivates the hexapod to adopt a tripod gait on various terrains. More information on rewards is detailed in Section \ref{style reward section}. These reward functions and their scales are listed in Table \ref{tab:reward details}.

\subsection{Gait-Rewards Induced Learning for Simple Tasks}

The primary task of the first stage is to enable a hexapod robot to perform velocity tracking tasks in a flat terrain environment using a tripod gait. The hexapod's design, featuring 18 joints across six legs, introduces significant redundancy that can disrupt training, often leading the robot to neglect two legs. Even when a tripod gait is achieved, the leg trajectories might not be symmetrical. To address this, we designed gait-related reward functions inspired from \cite{margolis2023walk} to effectively induce the robot to produce natural velocity tracking movements.

\textbf{Gait-Related Rewards:} In this stage, apart from the task-related reward \(r^g_t\), and the regularization reward \(r^l_t\), we established four types of gait-related rewards to regulate the robot's gait. The phase tracking function utilizes the difference between foot forces and velocities and the ideal swing-support state to induce a tripod gait. The Raibert Heuristic function calculates the desired foot position on the ground plane, adjusting the baseline stance width in line with the desired contact schedule and body velocity. The foot-swing height tracking function first computes each foot's desired contact state based on phase and timing variables, then calculates a penalty function based on the target foot height difference to constrain foot motion.

\textbf{Network architecture and Training:} The Stage I policy \({\pi _\theta^{\rm stage I}}\) comprises a low-level actor network and a critic network, both featuring the same architectural design. Their input, the proprioceptive observations $\boldsymbol{o}_t^p \in {\mathbb{R}^{60}}$, is processed through hidden layers of [128, 128, 64] dimensions and is directly trained using the PPO algorithm.

\textbf{Experiences Generation:} After training, the controller directs the robot in basic tasks like forward/backward movements, side steps, and turns. These actions are recorded as the robot performs a stable tripod gait, creating a 9.6-second trajectory experience dataset, which is then used for imitation learning in complex environments. Each state in dataset $\boldsymbol{s}_t^{AMP}$ in $\mathbb{R}^{42}$ includes joint positions, velocity, base linear and angular velocities. State transitions from the dataset $\mathcal{D}$ are used as real samples to train the discriminator. 

\subsection{Experience-Reward Induced Learning for Tough Tasks}
In the second stage, where complex environments may cause sudden gait changes, we address the challenge of effectively constraining movements through the Adversarial Motion Priors method, aiming to emulate biological progressive learning by drawing from previously accumulated motion experiences to generate more natural and robust movements. This method employs a GAN network to assess the similarity between current movements and reference experience trajectories, thereby generating a experience-guided reward signal that ensures the robot's natural and robust gait. 

\textbf{Experience-Guided Rewards:} \label{style reward section}In this stage, the reward function is composed of three elements: a task-related reward \(r^g_t\), a experience-guided reward \(r^e_t\), and a regularization reward \(r^l_t\), combined as $ \label{eq:rewardDefinition}r_t = r^g_t + r^e_t + r^l_t$. The experience reward assesses how closely the agent's actions mirror those of the demonstrator, with higher rewards for greater similarity. Given the superior stability of the tripod gait for hexapods on uneven terrains, we employ a experience-guided reward based on adversarial motion priors to encourage our robot to adopt a tripod gait, mirroring behaviors from a reference experience dataset \(\mathcal{D}\). Adopting the approach from \cite{peng2021amp}, we introduce a discriminator \(D_{\varphi}\), represented by a neural network with parameters \(\varphi\), to discern whether a state transition \({T}_s = (\boldsymbol{s}_t, \boldsymbol{s}_{t+1})\) is an authentic sample from \(\mathcal{D}\) or a fabricated sample by the policy \(\pi\). The discriminator's training objective is defined as: 
\begin{equation} \label{discriminator}
\begin{split}
&\quad\quad\quad\quad\quad\quad\quad\mathop{\arg\min} \limits_{\varphi} \mathcal{L}_1 +  \mathcal{L}_2  \\
\mathcal{L}_1 &= \mathbb{E}_{ {T}_s \sim \mathcal{D}} \left[\left(D_{\varphi}  ({T}_s)  - 1\right)^2\right] 
+\mathbb{E}_{  {T}_s \sim \pi} \left[\left(D_{\varphi}  ({T}_s) + 1\right)^2\right] \\
 \mathcal{L}_2 &= \frac{{{\alpha ^{gp}}}}{2}{_{{T}_s\sim{\cal D}}}\left[ \left \Vert {{\nabla _\varphi }{D_\varphi } ({T}_s)} \right\Vert_2 \right],
\end{split}
\end{equation}
where the first loss function $\mathcal{L}_1$ uses a least square GAN formulation, focusing on reducing the Pearson divergence between the distribution of the agent's state transitions and that of the reference data. This aims to train the discriminator to effectively identify whether a state transition originates from the policy $\pi$ or the reference experience dataset \(\mathcal{D}\). Additionally, we incorporate a gradient penalty in the second loss term $\mathcal{L}_2$ in Eq. (\ref{discriminator}) to prevent the discriminator from assigning non-zero gradients to the real data samples' manifold. This penalty is vital for ensuring stable training and effective performance, as demonstrated in \cite{peng2021amp}. The coefficient \(\alpha^{gp}\) is determined manually. The tripod style reward is then established based on:
\begin{equation} \label{style reward}
r_t^s\left[{T}_s\sim \pi\right] = \max \left[0, 1-0.25\left({D_\varphi } ({T}_s)- 1\right)^2 \right],
\end{equation}
where the experience-guided reward is scaled to the range $\left[0, 1\right]$.

\begin{table}[htbp]
\centering
\caption{dynamic parameters and the range of their randomization values used during training.}
\label{tab:random param}
\begin{tabular}{@{}lll@{}}
\toprule
\textbf{Parameters}& \textbf{Range[Min, Max]}         & \textbf{Unit}\\ \midrule
Link Mass       & {[}0.8, 1.2{]}$\times$nominal value  &  Kg          \\
Payload Mass    & {[}0, 5{]}                           &  Kg          \\
Payload Position& {[}-0.1, 0.1{]}relative to base position &  m       \\
Ground Friction & {[}0.05, 2.75{]}                     &  -          \\
Motor Strength  & {[}0.8, 1.2{]}                       &  -          \\
Joint $K_p$     & {[}0.8, 1.2{]}$\times$80             &  -          \\
Joint $K_d$     & {[}0.8, 1.2{]}$\times$1              &  -           \\
Joint Position  & {[}0.5, 1.5{]}$\times$nominal value  &  rad         \\ \bottomrule
\end{tabular}
\end{table}

\begin{table}[htbp]
\centering
\caption{Terrain types and the range of their level-Properties used during training.}
\label{tab:terrains type}
\begin{tabular}{@{}llll@{}}
\toprule
\textbf{Types} & \textbf{Level-Properties} & \textbf{Range{[}Min, Max{]}} & \textbf{Unit} \\ \midrule
Slopes (rough/normal)  & Slope inclination         & {[}0, 25{]}                 & deg           \\
Stairs (up/down)       & Step Height               & {[}0.05, 0.2{]}              & m             \\
Waves                  & Wave Amplitude            & {[}0.2, 0.5{]}               & m             \\
Discrete Steps         & $h^{\rm step}$            & {[}0.05, 0.15{]}             & m             \\ \bottomrule
\end{tabular}
\end{table}

\begin{table}[htbp]
\centering
\caption{Network architecture for two stages' policy and student policy. All networks use elu activations for hidden layers.}
\label{tab:net_arc}
\begin{tabular}{@{}lllll@{}}
\toprule
\textbf{Module} & \textbf{Inputs} & \textbf{Hidden Layers} & \textbf{Outputs} \\ \midrule
I Low-Level (MLP) & $o_t^p$   & {[}128, 128, 64{]}     &$a_t$  \\
I Critic (MLP)    & $o_t^p$          & {[}128, 256, 128{]}      & $V_t$  \\
II Low-Level (MLP) & $l_t,o_t^p$   & {[}256, 128, 64{]}     &$a_t$  \\
II Critic (MLP)    & $x_t$          & {[}512, 256, 128{]}      & $V_t$  \\
Memory (LSTM)   & $o_t^p,h_{t-1},c_{t-1}$& {[}256, 256, 256{]}      & $m_t$ \\
$g_p$ (MLP)& $s_t^p$    & {[}64, 32{]}            & $l_t^p$  \\
$g_e$ (MLP)& $i_t^e$   & {[}256, 128{]}          & $l_t^e$ \\
$g_m$ (MLP)& $m_t$ & {[}256, 128{]}& $l_t^{\rm student}$ \\
$D_{\varphi}$ (MLP)& $s_t^{AMP},s_{t+1}^{AMP}$ & {[}1024, 512{]}& $d_t^{\rm score}$ \\ \bottomrule
\end{tabular}
\end{table}

\textbf{Curriculum Design:} Training legged robots for blind locomotion on varied terrains involves significant challenges due to uncertain environmental interactions. Drawing on previous findings that diverse terrain training enhances complex locomotion skills, we introduce six types of procedurally generated terrains: slopes (both normal and rough), ascending and descending stairs, waves, and discrete steps. Details of terrain types and their difficulty ranges are provided in Table \ref{tab:terrains type}. Each terrain type is categorized into ten difficulty levels, with the rough slopes featuring added noise and the stairs having a consistent width. To foster omnidirectional navigational skills, we arrange slopes, large steps, and stairs in a pyramid formation, inspired by similar approaches in prior research. Given the initial instability of RL training, we employ a progressive curriculum, gradually introducing more complex terrains as the robot adapts to current levels, measured by its ability to maintain high linear velocity tracking rewards. Once a robot masters the highest terrain level, we cycle it back to a random level within the same terrain type and switch to a constant yaw command, promoting its ability to traverse complex terrains more effectively.

\textbf{Domain Randomization} To enhance our policy's robustness and ease its adaptation from simulations to real-world conditions, we vary several dynamics parameters in each episode which are outlined in Table \ref{tab:random param}

\textbf{Network architecture:}  The stage II policy \({\pi _\theta^{\rm stage II}}\) is composed of three parts: a terrain encoder \(g_e\), a privileged encoder \(g_p\), and a low-level network. The terrain encoder compresses terrain information $\boldsymbol{i}_t^e \in {\mathbb{R}^{187}}$ into a 16-dimensional latent space, while the privileged encoder reduces the privileged state $\boldsymbol{s}_t^p \in {\mathbb{R}^{42}}$ to an 8-dimensional latent representation. These encodings, combined with proprioceptive observations $\boldsymbol{o}_t^p \in {\mathbb{R}^{60}}$, are processed by the low-level network with a $tanh$ output layer to produce actions. Additionally, the policy includes a critic network presented by the MLP with three hidden layers for calculating target values in the advantage estimation. The discriminator \(D_{\varphi}\) is a simpler network with two hidden layers and a linear output. More details on each layer are shown in Table \ref{tab:net_arc}.

\textbf{Training:} We train the stage II policy using Proximal Policy Optimization (PPO) with access to privileged and terrain information.  Training of the policy and the discriminator occurs in synchronized. The policy generates state transitions $\boldsymbol{T}_{s}^{AMP}=\left(\boldsymbol{s}_t^{AMP},\boldsymbol{s}_{t+1}^{AMP}\right)$ for the discriminator $D_{\varphi}$ to evaluate ${D_\varphi } ({T}_s)$, contributing to the calculation of the style reward $r^e_t$. This stage's policy parameters $\theta$  are optimized for maximum return, while the parameters  $\varphi$  are fine-tuned to distinguish between real and agent-generated transitions.

\subsection{Deploy Training Based on Teacher-Student Methods}

Due to the lack of exteroceptive sensory input in physical world, the terrains remain only partially observed, rendering the blind locomotion scenario a Partially Observable Markov Decision Process (POMDP).To realize the deployment of trained agent in the real world, we utilize a method known as privileged learning, as explored by \cite{chen2020learning}.  
The 'teacher' policy, referring to the stage II policy, is distilled through supervised learning into a 'student' policy. This 'student' policy is trained to infer dynamic characteristics from a sequence of past observations, effectively embodying the knowledge and strategies of the stage II policy.

\textbf{Network architecture:} The student policy is built with a memory encoder and an MLP, identical in structure to the teacher's low-level net. We chose an LSTM-based RNN, which efficiently embeds historical information in its hidden states. Here, proprioceptive observations \(o_t^p\) and previous states \((h_{t-1}, c_{t-1})\) are encoded by the RNN into intermediate states \(m_t\), and then processed by a neural network \(g_m\) to produce the student's latent representation \(l_t^s\). To accelerate training, the student's low-level net is initialized with the teacher's pretrained weights. More details are in Table \ref{tab:net_arc}.

\textbf{Training:} The student policy is trained to replicate the teacher's actions, operating without privileged state \(\boldsymbol{s}_t^p\) or terrain information \(\boldsymbol{i}_t^e\). This creates a Partially Observable Markov Decision Process (POMDP), where the student must use observation history \(\boldsymbol{o}_t^p\) to infer unobservable states. The student's memory encoder is responsible for understanding the sequential relationship between these histories. Training involves two losses: imitation and reconstruction, the former for action mimicry and the latter for replicating the teacher's latent representations. We adopt the Dataset Aggregation (DAgger) strategy for robustness, to generate samples by rolling out the student policy. The student undergoes the same terrain curriculum as the teacher, but without a discriminator.

\section{EXPERIMENTAL SETUP}
\textbf{Simulation:} In our training, we simultaneously engaged 4096 agents across 30,000 episodes. This comprised 5000 episodes for the Stage I policy, 15,000 episodes for the staget II policy and 10,000 for the student policy, with the training conducted in diverse terrains using the IsaacGym simulator\cite{rudin2022learning}. Each RL episode was capped at 1000 steps, equating to 20 seconds, with early termination possible upon meeting specific criteria. The policies operated at a control frequency of 50 Hz, with each simulation step representing 0.02 seconds. All training costs about 20 hours on a single NVIDIA RTX 4090 GPU. The training of the hexapod and quadruped robots employ the same setup, being validated in the Gazebo simulation environment.

\textbf{Hardware:} We implemented our controller on the Unitree Go1 Edu robot, measuring 0.3 meters in height and weighing 13 kilograms. The robot is equipped with joint position encoders and an IMU as its primary sensors. Our trained policy operates on the robot's onboard Jetson TX2 NX computer, executing control commands at a frequency of 50 Hz..

 \section{RESULTS AND DISCUSSION}

\subsection{Ablation Study for Experience-Reward}
\begin{figure}[htbp]
\includegraphics[width=\linewidth]{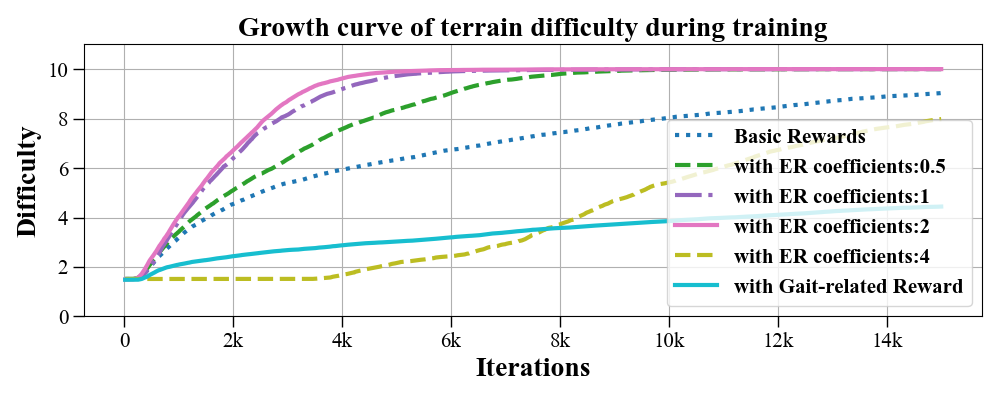}
\label{fig:terrain difficulty}
\centering
\caption{The variation in terrain difficulty during training with the same random seed under different rewards indicates the robot's learning speed for effective motions. Basic rewards combined with well-scaled Experience-Guided rewards enhance motion sampling. However, manually set gait rewards hinder effective learning, leading to minimal increases in terrain difficulty. This demonstrates the effectiveness of Experience-Guided rewards in improving learning in complex terrains.}
\end{figure}

We performed ablation experiments with a hexapod robot for velocity tracking in complex environments, including: (a) training with only basic task rewards $r^g_t$ and regularization rewards $r^l_t$; (b) adding gait reward $r^{gait}_t$ to basic rewards; (c) using experience-guided reward $r^e_t$ over basic rewards, varying coefficients to evaluate training impact. Basic reward coefficients were constant, as detailed in Table \ref{tab:reward details}. We assessed reward function effectiveness by analyzing terrain difficulty trend curves under various settings (Fig \ref{fig:terrain difficulty}), where terrain difficulty rises with significant reward achievement. Higher terrain difficulty growth rates indicate more effective rewards. Basic rewards alone led to some learning of traversable motions, but these were often unnatural due to the large search space. Incorporating Experience-Guided rewards quickened terrain difficulty escalation, hinting at more efficient motion learning. However, very high coefficients of this reward reduced learning efficiency, suggesting a balance is needed in mimicking flat terrain movements for complex environments. Basic rewards plus manually set gait rewards struggled with adapting to terrain changes, thus limiting movement learning and terrain difficulty progression.

\begin{figure}[]
\includegraphics[width=\linewidth]{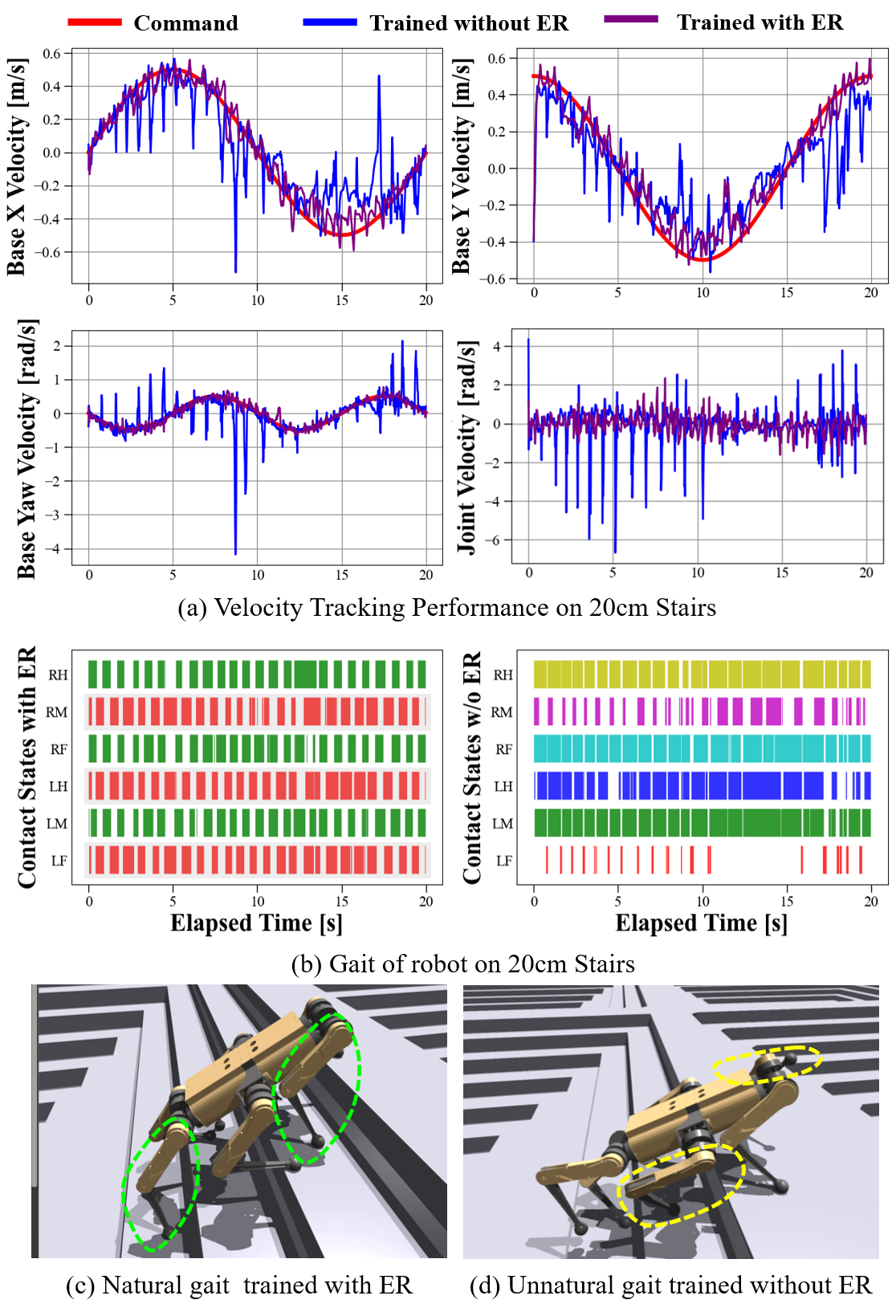}

\label{fig:sin track}
\centering
\caption{Comparison of Velocity Tracking Performance and Gait on 20cm Stairs: Evaluating Robot Control with Policies Trained Using Experience-Guided Rewards (ER) Versus Without.}
\end{figure}

\subsection{Evaluation of the Natural and Robust Locomotion}

After training, the most challenging 20cm staircase was used as a test site to verify the effectiveness of the experience-guided reward function, with the velocity tracking performances and gait behaviors showcased in Fig. \ref{fig:sin track}. The robot received various sine velocity commands ($V_x$, $V_y$, $W_z$) with different frequencies and amplitudes to assess its velocity tracking robustness and the naturalness of its gait. Training with added gait rewards failed to produce effective obstacle-crossing gaits, as direct gait rewards overly constrained the robot's movements, hindering effective exploration and sampling in complex environments. Hence, we didn't present its movement results. Basic task and regularization rewards generated gaits with velocity tracking capability, but these were unstable and irregular . Additionally, its z-direction speed, joint torques, and velocities were more oscillatory. Foot contact forces, shown in the right chart of Fig. \ref{fig:sin track} (b), displayed an irregular gait with legs RH, RF, LH, LM contacting the ground for extended periods while RM and LF barely touched the ground, leading to an unnatural movement captured in Fig. \ref{fig:sin track} (d). The robot's body was low, and the irregular swinging of legs RM and LF (yellow circle) resembled a bound-like irregular gait. 

In contrast, the inclusion of experience-guided reward signals resulted in a natural and robust diagonal gait, as shown in Fig. \ref{fig:sin track} (a) (purple curve),, where the robot tracked velocity commands with minimal error, even on 20cm high stairs. The robot's joint torques and velocities during movement were more stable without additional rewards. We believe the experience reward \(r^e_t\) enabled the strategy to learn behaviors capturing the essence of the reference tripod gait, allowing the robot to autonomously learn a tripod gait in complex environments similar to flat terrain. As seen in the left chart of Fig. \ref{fig:sin track} (b), legs LF, LH, RM moved in nearly identical phases, with the remaining diagonal legs similarly synchronized. The robot autonomously adjusted its step frequency to navigate complex terrains without breaking the tripod gait, as illustrated in Fig. \ref{fig:sin track} (c), where legs RF and RH (green labels) moved in almost identical states, crossing obstacles with a robust gait.

\subsection{Hardware Testing Performance}
\begin{figure}[]
\includegraphics[width=\linewidth]{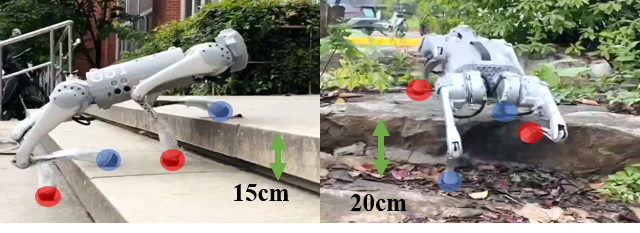}
\label{fig:go1}
\centering
\caption{Naturally Robust trot gait in physic robot Go1. The blue icon indicates the support phase and the red icon the swing phase, demonstrating the robot's consistent diagonal gait on stairs of varying heights.}
\end{figure}

To evaluate the hardware robustness of our training, we used the Unitree-Go1 as a test platform, successfully transferring our policy using a teacher-student strategy. This led to natural, robust gaits in complex terrains. We compared our method against basic rewards (BR), BR with gait-rewards (BR+GR), and BR with experience-reward (BR+ER) on a 20cm high staircase. After conducting five trial sets with differently trained network controllers, our method consistently achieved a 100\% success rate in climbing the stairs, maintaining natural and robust gait, is shown in Fig \ref{fig:go1}.

\begin{table}[htbp]
\centering
\caption{Success rates of different methods for different step heights}
\begin{tabular}{cccc}
\hline
\textbf{Methods} & \textbf{BR} & \textbf{BR+GR} & \textbf{BR+ER}\\
\hline
{\textbf{10cm}} & 60\% & 20\% & 100\% \\
{\textbf{15cm}} & 40\% & 0\% & 100\% \\
{\textbf{20cm}} & 40\% & 0\% & 100\% \\
\hline
\end{tabular}
\label{tab:measurements}
\end{table}

\section{CONCLUSIONS}

In this study, we introduce a bioinspired two-stage learning framework with two-step reward that efficiently enables diverse legged robots to learn naturally robust movements in complex settings. Starting with manual reward function adjustments for natural gait generation on flat terrain, we then leverage biological learning principles, using these gaits as a baseline for more complex task learning. This method not only minimizes the need for extensive manual tuning but also circumvents the challenges of deriving optimized movement patterns through model analysis or animal motion capture. Applicable to a wide range of legged robots, including those with varying scales and rigid-flexible coupling, this framework can also be extended to robotic arms and other robots. Our future research will focus on identifying the most beneficial experiences, devising strategies for their effective integration, and exploring the potential for autonomous selection of motion priors by robots for enhanced learning, potentially revealing secrets of biological motion learning.

\bibliographystyle{./bibtex/IEEEtran} 
\bibliography{./bibtex/my_refs}

\end{document}